\documentclass[nohyperref]{article}

\usepackage{microtype}
\usepackage{graphicx}
\usepackage{subcaption}
\usepackage{booktabs} 

\usepackage{hyperref}

\PassOptionsToPackage{dvipsnames}{xcolor}

\usepackage[accepted]{icml2022}

\usepackage{amsmath}
\usepackage{amssymb}
\usepackage{mathtools}
\usepackage{amsthm}

\usepackage{mathrsfs}
\usepackage{commath}

\usepackage[capitalize,noabbrev]{cleveref}

\theoremstyle{plain}
\newtheorem{theorem}{Theorem}[section]
\newtheorem{proposition}[theorem]{Proposition}

\theoremstyle{definition}
\newtheorem{definition}[theorem]{Definition}

\theoremstyle{remark}

\usepackage[textsize=tiny]{todonotes}

\newcommand{\T}{\ensuremath{\mathsf{T}}}

\icmltitlerunning{Fenrir: Physics-Enhanced Regression for Initial Value Problems}

\begin{document}

\twocolumn[
\icmltitle{Fenrir: Physics-Enhanced Regression for Initial Value Problems}



\icmlsetsymbol{equal}{*}

\begin{icmlauthorlist}
\icmlauthor{Filip Tronarp}{equal,yyy}
\icmlauthor{Nathanael Bosch}{equal,yyy}
\icmlauthor{Philipp Hennig}{yyy,comp}
\end{icmlauthorlist}

\icmlaffiliation{yyy}{Department of Computer Science, University of T\"ubingen, T\"ubingen, Germany}
\icmlaffiliation{comp}{Max--Planck Institute for Intelligent Systems, T\"ubingen, Germany}

\icmlcorrespondingauthor{Filip Tronarp}{filip.tronarp@uni-tuebingen.de}
\icmlcorrespondingauthor{Nathanael Bosch}{nathanael.bosch@uni-tuebingen.de}

\icmlkeywords{Machine Learning, ICML}

\vskip 0.3in
]



\printAffiliationsAndNotice{\icmlEqualContribution} 

\begin{abstract}
We show how probabilistic numerics can be used to convert an initial value problem into a Gauss--Markov process parametrised by the dynamics of the initial value problem.
Consequently, the often difficult problem of parameter estimation in ordinary differential equations is reduced to hyperparameter estimation in Gauss--Markov regression, which tends to be considerably easier.
The method's relation and benefits in comparison to classical numerical integration and gradient matching approaches is elucidated.
In particular, the method can, in contrast to gradient matching, handle partial observations, and has certain routes for escaping local optima not available to classical numerical integration.
Experimental results demonstrate that the method is on par or moderately better than competing approaches.
\end{abstract}

\section{Introduction}
\label{sec:introduction}
Consider the following initial value problem (IVP)
\begin{equation}\label{eq:ivp}
  \frac{\dif}{\dif t} \varphi_{\theta}(t) = f_\theta \left( t, \varphi_\theta(t) \right),
  \qquad t \in [0,T],
\end{equation}
where the vector field $f_\theta \colon [0,T] \times \mathbb{R}^d \to \mathbb{R}^d$ and the initial condition \(\varphi_\theta(0) = y_0(\theta)\) are both parametrised by $\theta$. 
In this article, the concern lies in estimating $\theta$ from noisy measurements of the following form
\begin{equation}\label{eq:measurements}
u(t) = H^\T \varphi_\theta(t) + v(t), \quad v(t) \sim \mathcal{N}(0,R_\theta),
\end{equation}
where $t \in \mathbb{T}_{\mathsf{D}} \subset [0,T]$ is the finite set of measurement nodes and
\(H\) is a measurement matrix of appropriate dimension.
This is a ubiquitous problem in science and engineering.
Examples include ecology \citep{Benson1979}, pharmacokinetics \citep{Gelman1996}, process engineering \citep{Astrom1971}, and brain imaging \citep{Friston2002}.

The likelihood functional $\mathcal{L}_{\mathsf{D}}$, evaluated at some function $y$, is given by
\begin{equation*}
\mathcal{L}_{\mathsf{D}}(R_\theta,y) = \prod_{t \in \mathbb{T}_{\mathsf{D}} } \mathcal{N} \left( u(t);H^\T y(t),R_\theta \right),
\end{equation*}
and the marginal likelihood $\mathcal{M}$ of some parameter $\theta$ can be expressed by evaluating $\mathcal{L}_{\mathsf{D}}$ at the corresponding solution $\varphi_\theta$ according to
\begin{equation*}
\mathcal{M}(\theta) = \mathcal{L}_{\mathsf{D}} \left( R_\theta, \varphi_{\theta} \right).
\end{equation*}
The parameter $\theta$ may be estimated by maximising $\mathcal{M}$.
A persistent challenge in likelihood-based inference in initial value problems is the fact that $\varphi_{\theta}$, and therefore the likelihood, are intractable \citep{Bard1974}.

\begin{figure}[t]
  \centering
  \includegraphics{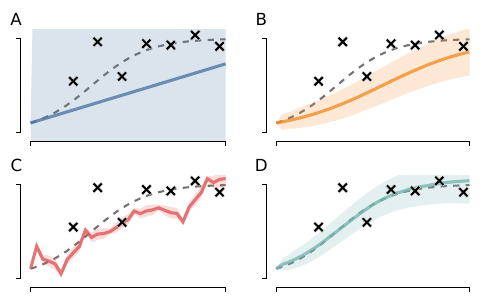}
  \caption{
    \textbf{From an uninformed prior to a calibrated posterior.}
    Starting from a standard Gauss--Markov prior (A), Fenrir first computes a physics-enhanced prior with probabilistic numerics (B), and then a posterior via Gauss--Markov regression (C).
    By maximizing the marginal likelihood, we obtain a calibrated posterior and parameter estimates for the underlying dynamical system (D).
    The data generating model is the logistic equation, which corresponds to the vector field $f(t,y) = ry(1-y)$.
  }
  \label{fig:1}
\end{figure}

A standard appproach to approximating the likelihood is based on solving the IVP numerically \citep{Hairer87}.
However, in optimisation-based inference it has been observed that this leads to many local optima \citep{Cao2011}, and can lead to divergence of the optimiser \citep{Dass2017}.
On the other hand, slow convergence and poor mixing has been observed for Monte Carlo-based inference \citep{Alahmadi2020}, which have led some authors to favour likelihood-free methods \citep{Toni2009}.
Another alternative is gradient matching \citep{Voit2000}
with splines
\citep{Varah1982,Gugushvili2012}
or Gaussian processes
\citep{Calderhead2009,Dondelinger2013,Gorbach2017,Wenk2020}.

\subsection{Contribution}
In the present work, a probabilistic numerics approach is developed for computing the marginal likelihood.
Probabilistic numerics aims at producing probability measures for solutions of numerical problems,
thus giving a probabilistic description of the numerical error \citep{Hennig2015,Oates2019a}.

The marginal likelihood may be viewed as $\mathcal{L}_{\mathsf{D}}$ integrated against a Dirac measure located at $\varphi_{\theta}$ according to
\begin{equation}\label{eq:exact_marginal_likelihood}
\mathcal{M}(\theta) = \int \mathcal{L}_{\mathsf{D}}(R_\theta,y) \delta (y - \varphi_{\theta} ) \dif y.
\end{equation}
While this representation is not immediately advantageous, it is instructive for understanding the probabilistic numerics approach.
Namely, it produces an approximation to the Dirac measure, giving the following approximate marginal likelihood
\begin{equation}\label{eq:approximate_marginal_likelihood}
\widehat{\mathcal{M}}_N(\theta,\kappa) = \int \mathcal{L}_{\mathsf{D}}(R_\theta,y)  \widehat{\delta}_N (y \mid \theta, \kappa) \dif y,
\end{equation}
where $\widehat{\delta}_N$ is the output of a suitably chosen probabilistic numerical method, which is parametrised by $\kappa$.
It should be noted that $\mathcal{L}_{\mathsf{D}}$ only depends on point evaluations of $y$ on the grid $\mathbb{T}_{\mathsf{D}}$.
Therefore, it is sufficient to operate on the finite dimensional distributions of $\widehat{\delta}_N$ to compute $\widehat{\mathcal{M}}_N$.

\citet{Kersting2020b} has previously used the representation \eqref{eq:approximate_marginal_likelihood} and approximated its gradients in combination with low order explicit solvers, at a cost of $O(N^3)$.

The aim of this article is to show how both $\widehat{\delta}_N$ and $\widehat{\mathcal{M}}_N$ can be computed efficiently for general probabilistic solvers, at a cost of $O(N)$.
The method consists of two parts:
\begin{enumerate}
\item Efficiently construct a Gauss--Markov representation of $\widehat{\delta}_N (y \mid \theta,\kappa)$ using probabilistic numerics.
\item Compute $\widehat{\mathcal{M}}_N(\theta,\kappa)$ and its derivatives
  via Gauss--Markov regression and automatic differentiation.
\end{enumerate}

The first step essentially takes the initial value problem and produces a \emph{physics-enhanced} Gauss--Markov prior.
The second step utilises this prior in standard Gauss--Markov regression to estimate parameters and reconstruct the trajectory \citep{Sarkka2019}.
Therefore, the method is called \underline{Ph}ysics-\underline{en}hanced \underline{r}egression in \underline{i}nitial value p\underline{r}oblems, or \emph{Fenrir} for short.
Here physics is used to refer to any mechanistic information pertaining to the dynamics of the data generating process.
The method is illustrated in \cref{fig:1}.

The rest of the article is organised as follows.
Probabilistic numerical solvers are reviewed in \cref{sec:pn}.
In \cref{sec:fenrir} it is shown how to use probablistic numerics to construct a physics-enhanced Gauss--Markov prior for initial value problems,
thus reducing the marginal likelihood to Gauss--Markov regression.
Related work is discussed in \cref{sec:related}, which is followed by experimental results in \cref{sec:experiments}.
Finally, concluding remarks are given in \cref{sec:conclusion}.

\section{Probabilistic Numerical IVP Solvers}\label{sec:pn}
In the Bayesian formulation, an IVP solver is completely specified by a prior and the definition of the data, on which it is conditioned.
The latter is obtained by means of an information operator \citep{Cockayne2019a}.
For constructing a probabilistic numerical solver, we follow the account of \citet{Tronarp2019c,Tronarp2021a}.

\subsection{Prior Specification}
The probabilistic numerics prior is defined as the output of the following stochastic state-space model
\begin{subequations}\label{eq:pn_prior}
\begin{align}
\dif x(t) &= A x(t) \dif t + \sqrt{\kappa} B \dif w(t), \quad x(0) = x^\dagger_\theta,\\
y^{(m)}(t) &= \mathrm{E}_m^\T x(t), \quad m = 0,1,\ldots,\nu,
\end{align}
\end{subequations}
where $x \in \mathbb{R}^{d(\nu+1)}$ models the solution and its $\nu$ first derivatives and $\mathrm{E}_m$ are selection matrices for the $m$th derivative of the prior model for the solution of \eqref{eq:ivp}, which is denoted by $y$.
Furthermore, $x^\dagger_\theta$ denotes the initial condition of $x$, $A \in \mathbb{R}^{d(\nu+1)\times d(\nu+1)}$ and $B \in \mathbb{R}^{d(\nu+1)}$ are model matrices and $w$ is a standard Wiener process in $\mathbb{R}^d$ \citep{Oksendal2003}.

The state $x$ is a Markov process by construction, with transition density given by \citep{Sarkka2019}
\begin{align*}
\Phi(h) &=  e^{Ah}, \\
Q(h)    &= \int_0^h \Phi(h-\tau) BB^\T \Phi^\T(h-\tau) \dif \tau, \\
x(t+h) &\mid x(t) \sim \mathcal{N}\big(x(t+h); \Phi(h)x(t), \kappa Q(h)\big),
\end{align*}
which facilitates fast computation for the probabilistic solver and our subsequent marginal likelihood approximation.
Additional details on priors for probabilistic solutions of initial value problems can be found in \cref{app:pnprior}.

\subsection{Data Model}
In order to define a data model for probabilistic numerical solvers, a grid
\begin{equation*}
\mathbb{T}_{\mathsf{PN}} = \{t_n\}_{n=1}^N \subset [0,T],
\end{equation*}
needs to be coupled with an information operator. The canonical information operator for initial value problems is given by \citep{Tronarp2021a}
\begin{equation}\label{eq:pn_data_model}
\mathcal{Z}_\theta[x](t) = \mathrm{E}_1^\T x(t) - f_\theta(t,\mathrm{E}_0^\T x(t)),
\end{equation}
but there are alternatives that, for instance, also take geometric invariants into account \citep{Bosch2021b}.

Note that $\mathcal{Z}$ map solutions of the initial value problem to the zero function, which is a known value.
In fact, the set of functions starting at $y_0(\theta)$ which are mapped to the zero function by $\mathcal{Z}$ constitutes the set of solutions to the initial value problem \cite{Arnold1992}.%
\footnote{Typically, we assume that the vector field is regular enough for there to be a unique solution of the initial value problem.}
An appropriate data model for a probabilistic numerical solver is thus given by
\begin{equation}\label{eq:pn_data}
z(t) = \mathcal{Z}_\theta[x](t) = 0, \quad  t \in \mathbb{T}_{\mathsf{PN}},
\end{equation}
where $z(t) = 0$ is enforced only on the chosen grid as to arrive at a practical algorithm.

It should be noted that the grid $\mathbb{T}_{\mathsf{PN}}$ does not need to be specified a priori but can be constructed adaptively to control the solution error \citep{Schober2019,Bosch2021}.

\subsection{Initial Value Problem Solvers as Non-linear Gauss--Markov Regression}
The prior \eqref{eq:pn_prior}, data model \eqref{eq:pn_data_model}, and data definition \eqref{eq:pn_data} define a non-linear Gauss--Markov regression problem according to \citet{Tronarp2019c}
\begin{subequations}\label{eq:ivp2gmr}
\begin{align}
x(t_n) \mid x(t_{n-1})   &\sim \mathcal{N}\big(\Phi(\Delta_n)x(t_{n-1}), \kappa Q(\Delta_n)\big), \\
z(t_n) \mid x(t_n)       &\sim \mathcal{N}\big( \mathrm{E}_1^\T x(t) - f_\theta(t,\mathrm{E}_0^\T x(t)), 0  \big), \\
z(t_n)  &\coloneqq 0,
\end{align}
\end{subequations}
where $\Delta_n = t_n - t_{n-1}$ is the step-size of the $n$th step, $x(t_0) = x^\dagger_\theta$ by convention, and $\mathcal{N}(\cdotp,0)$ denotes the Dirac distribution.
The probablistic numerical solver for \eqref{eq:ivp} associated with the prior \eqref{eq:pn_prior} and the data \eqref{eq:pn_data} is on the grid $\mathbb{T}_{\mathsf{PN}}$ given by
\begin{equation}\label{eq:pn_posterior}
\begin{split}
&\gamma_N( t_{1:N}, x_{1:N} \mid \theta,\kappa ) =  c^{-1}(\theta,\kappa) \\
&\quad \times \prod_{n=1}^N \mathcal{N}\big( x_n ; \Phi(\Delta_n) x_{n-1}, \kappa Q(\Delta_n) \big) \\
&\quad \times \prod_{n=1}^N \delta \big( \mathrm{E}_1^\T x_n - f_\theta( t_n, \mathrm{E}_0^\T x_n )  \big),
\end{split}
\end{equation}
where $c(\theta,\kappa)$ is a norming constant. Due to the potential non-linearity of the vector field, this object is generally intractable.
However, when the vector field is linear, say
\begin{equation}\label{eq:affine_vector_field}
f_\theta(t,y) = L_\theta(t) y + b_\theta(t),
\end{equation}
then the densities of the time marginals can be computed efficiently via Kalman filtering and Rauch--Tung--Striebel smoothing \citep{Kalman1960,RauchTungStriebel1965}.

This fact is exploited for approximate inference when the vector field is non-linear as well. Indeed several linearisation approaches have been employed \citep{Schober2019,Tronarp2019c,Tronarp2021a},
which have been demonstrated to yield accurate solvers both empirically  \citep{Schober2019,Bosch2021,Kramer2020a} and theoretically \citep{Kersting2020a,Tronarp2021a}.

\subsection{Initial Value Problem Solvers as Kalman Filtering}
The Kalman filtering recursion for \eqref{eq:ivp2gmr}
when the vector field is affine as in \eqref{eq:affine_vector_field},
recursively computes the densities
\begin{equation}
\pi( x(t_n) \mid z(t_{1:n}) ) = \mathcal{N}( \mu_\theta(t_n), \Sigma_\theta(t_n) ),
\end{equation}
which are the time marginals conditioned on all past data up to the present.
The recursion is initialised by setting $\mu_\theta(t_0) = x^\dagger_\theta, \Sigma_\theta(t_0) = 0$, and then alternates between prediction and update.
\begin{itemize}
\item Prediction:
\begin{align*}
\mu_\theta(t_n^-)     &= \Phi(\Delta_n) \mu_\theta(t_{n-1}), \\
\Sigma_\theta(t_n^-)  &= \Phi(\Delta_n) \Sigma_\theta(t_{n-1}) \Phi^\T(\Delta_n) +  Q(\Delta_n).
\end{align*}
\item Update:
\begin{align*}
C_\theta(t_n) &= \mathrm{E}_1 - \mathrm{E}_0 L_\theta^\T(t_n), \\
S_\theta(t_n) &= C_\theta^\T(t_n) \Sigma_\theta(t_n^-) C_\theta(t_n), \\
K_\theta(t_n) &= \Sigma_\theta(t_n^-) C_\theta^\T(t_n) S_\theta^{-1}(t_n), \\
e_\theta(t_n)        &= b_\theta(t_n) - C_\theta^\T(t_n) \mu_\theta(t_n^-), \\
\mu_\theta(t_n) &= \mu_\theta(t_n^-) + K_\theta(t_n) e(t_n),\\
\Sigma_\theta(t_n) &= \Sigma_\theta(t_n^-) - K_\theta(t_n) S_\theta(t_n) K_\theta^\T(t_n).
\end{align*}
\end{itemize}
The following parameters can be  computed from the outputs of the Kalman filter
\begin{subequations}\label{eq:backwards_parameters}
\begin{align}
&G_\theta(t_{n-1}) = \Sigma_\theta(t_{n-1}) \Phi^\T(\Delta_n)\Sigma_\theta^{-1}(t_n^-), \\
&P_\theta(t_{n-1})     = \Sigma_\theta(t_{n-1}) - G_\theta(t_{n-1}) \Sigma_\theta(t_n^-) G_\theta^\T(t_{n-1}).
\end{align}
\end{subequations}
They are used for the smoothing recursion and the representation of the probabilistic numerics posterior.

\section{Fenrir}\label{sec:fenrir}
In this section, it is shown that the probabilistic numerical solver yields a Gauss--Markov process approximation to \eqref{eq:ivp}.
Consequently, inference given measurements \eqref{eq:measurements} reduces to a Gauss--Markov regression problem with a \emph{physics-enhanced prior} as determined by the probabilistic solver.

\subsection{Probabilistic Numerical IVP Solutions as Gauss--Markov Processes}
Linearising the vector field allows for approximate computation of the time marginal densities via the Rauch--Tung--Striebel smoother.
These linearisations imply a Gauss--Markov representation of the approximate posterior, which in fact is used in the Bayesian derivation of the smoothing algorithm \citep[c.f.\ proof of theorem 8.2]{Sarkka2013}.
The following result lie at the heart of our method.
\begin{proposition}[Gauss--Markov representation of the probabilistic solver]\label{prop:gauss_markov_pn_post}
The restriction of the probabilistic numerics posteriors to the grid $\mathbb{T}_{\mathsf{PN}}$ admit the following representation
\begin{equation}\label{eq:ivp2gm}
\begin{split}
&\widehat{\gamma}_N(t_{1:N},x_{1:N}\mid \theta, \kappa) = \mathcal{N} \big( x_N ; \xi_\theta(t_N), \kappa \Lambda_\theta(t_N)   \big) \\
&\quad \prod_{n=N-1}^1 \mathcal{N}\big( x_n ; G_\theta(t_n)  x_{n+1} + \zeta_\theta(t_n), \kappa P_\theta(t_n)   \big),
\end{split}
\end{equation}
where $\xi_\theta(t_N)=\mu_\theta(t_N)$, $\Lambda_\theta(t_N) = \Sigma_\theta(t_N)$,
\begin{equation*}
\zeta_\theta(t_n) = \mu_\theta(t_n) - G_\theta(t_n) \mu_\theta(t_{n+1}^-),
\end{equation*}
and $(G,P)$ are given by \eqref{eq:backwards_parameters}.
\end{proposition}
For completeness, a detailed derivation of proposition \ref{prop:gauss_markov_pn_post} is given in Appendix  \ref{app:pnlin}. Note that $\widehat{\gamma}_N$ is represented as a Gauss--Markov process running backwards in time.
It represents a probabilistic approximation to the solution of the IVP and its derivatives, in terms of a conditional distribution given numerical data \eqref{eq:pn_data} and the parameter $\theta$.

\subsection{Inference in IVPs as Gauss--Markov Regression}
In the previous section, the approximate Dirac $\widehat{\delta}_N(y\mid \theta)$ was implicitly defined through $\gamma_N$ in \eqref{eq:pn_posterior}.
The purpose here is to turn this into an implementable algorithm for approximating the marginal likelihood.
For ease of notation it is assumed that $\mathbb{T}_{\mathsf{D}} \subset \mathbb{T}_{\mathsf{PN}}$, in which case,
\begin{equation*}
\begin{split}
&\widehat{\mathcal{M}}_N(\theta,\kappa) = \int \mathcal{L}_{\mathsf{D}}(\theta,y)  \widehat{\delta}_N (y \mid \theta,\kappa) \dif y \\
&\quad= \int  \mathcal{L}_{\mathsf{D}}(R_\theta,\mathrm{E}_0^\T x) \gamma_N(t_{1:N},x_{1:N}\mid \theta, \kappa) \dif x_{1:N}.
\end{split}
\end{equation*}
Additionally, the calibration parameter $\kappa$ is also included in the marginal likelihood approximation.
In practice, $\gamma_N$ is replaced by its approximation $\widehat{\gamma}_N$ in \eqref{eq:ivp2gm}.
This results in the following approximation to the marginal likelihood
\begin{equation}\label{eq:mlapproximation}
\begin{split}
\widehat{\mathcal{M}}_N(\theta,\kappa) &=\int  \prod_{t_n \in \mathbb{T}_{\mathsf{D}} } \mathcal{N}(u(t_n);H^\T \mathrm{E}_0^\T x_n,R_\theta) \\
&\quad \times\widehat{\gamma}_N(t_{1:N},x_{1:N}\mid \theta, \kappa) \dif x_{1:N}.
\end{split}
\end{equation}
Consequently, the problem of computing the marginal likelihood and trajectory estimates is reduced to inference in the following linear state-space model
\begin{subequations}\label{eq:fenrir}
\begin{align}
x(t_N) &\sim \mathcal{N}( \xi(t_N), \kappa \Lambda(t_N) ), \\
x(t_n) \mid x(t_{n+1}) &\sim \widehat{\gamma}_N(x(t_n) \mid x(t_{n+1}), \theta, \kappa), \\
u(t) \mid x(t) &\sim \mathcal{N}(  H^\T \mathrm{E}_0^\T x(t), R_\theta ), \quad t \in \mathbb{T}_{\mathsf{D}}, \label{eq:measurements2}
\end{align}
\end{subequations}
where the backwards transition densities can be read from \eqref{eq:ivp2gm}. Therefore, estimating the trajectory of the solution \eqref{eq:ivp} can also be done via Kalman filtering and smoothing.
Furthermore, the marginal likelihood approximation can be computed via the Kalman filter through the prediction error decomposition \citep{Schweppe1965}.
Complete details on how to compute trajectory estimates and marginal likelihoods in \eqref{eq:fenrir} are given in \cref{app:ivp2gmr}.

\paragraph{Computational complexity}
The computation of $\widehat{\gamma}_N$ and $\widehat{\mathcal{M}}_N$ can be implemented with Gauss--Markov regression with a state dimension of $d(\nu+1)$.
Therefore, assuming the measurement dimension is smaller, the computational complexity of the method is $O( Nd^3(\nu+1)^3)$.
That is, it is linear in the number of data points, in contrast to cubic complexity for standard Gaussian process regresison.
Further speed-ups may be obtainable by exploiting structural simplifications for certain probabilistic solvers \citep{Kramer2021a}.

\paragraph{Hyperparameter estimation}
The present method provides a marginal likelihood \eqref{eq:mlapproximation}; its derivatives can be computed with automatic differentiation.
Consequently, Fenrir interacts with various inference methods, such as gradient-based optimisation or Markov Chain Monte Carlo, in a plug-and-play fashion.
In this paper, the maximum likelihood approach is examined.

\paragraph{Model selection}
The marginal likelihood approximation \eqref{eq:mlapproximation} confers other benefits than providing a cost function for parameter inference.
Namely, the possibility for a probabilistically motivated model comparisons,
such as likelihood ratio testing for nested models \citep{King1998}, or via various information criteria \citep{Akaike1974,Stoica2004}.

\section{Related Work: A Tale of Three Approaches}\label{sec:related}
Three different approaches to parameter estimation in initial value problems can be discerned, namely (a) numerical integration, (b) gradient matching, and (c) probabilistic numerics.
%
%
In  order to get a comprehensive lay of the land of parameter estimation in ordinary differnetial equations, these approaches are reviewed in this section.
Particular care is taken to highlighting similarities and differences.

\subsection{Classical Numerical Integration}
The traditional approach is to estimate the parameters via non-linear regression \citep{Biegler1986},
where the correct solution to \eqref{eq:ivp} is replaced by a numerical approximation, say Runge--Kutta \citep{Hairer87}.
Thus the marginal likelihood approximation reads
\begin{equation}\label{eq:rk_likelihood}
\begin{split}
\widehat{\mathcal{M}}_N(\theta) &=\int  \prod_{t_n \in \mathbb{T}_{\mathsf{D}} } \mathcal{N}(u(t_n);H^\T y_n,R_\theta) \\
&\quad \times\prod_{ t_n \in \mathbb{T}_{\mathsf{D}}  }  \delta(y_n - \hat{\varphi}_{\theta}(t_n) )   \dif y_{1:N}.
\end{split}
\end{equation}
That is, likelihood computation via numerical integration computes the Dirac approximation $\widehat{\delta}_N$ in \eqref{eq:approximate_marginal_likelihood}
by approximating the location of the Dirac in \eqref{eq:exact_marginal_likelihood}.

\subsection{Gradient Matching}
The main idea of gradient matching is to decompose the inference procedure into two steps:
\begin{enumerate}
\item Fit a curve $\hat{y}(t)$ to the data $u(t), t \in \mathbb{T}_{\mathsf{D}}$.
\item Estimate the parameter $\theta$ by minimising the deviation from the differential equation:
$\dot{\hat{y}}(t) - f_\theta(t,\hat{y}(t))$.
\end{enumerate}
This procedure is vaguely formulated, purposely so.
Indeed, different alternatives for these steps have surfaced throughout the years.

\paragraph{Spline smoothing} The first approach was to implement the curve fitting step with splines  \citep{Varah1982} or kernel regression \citep{Gugushvili2012},
whereafter the gradient matching step is posed as a non-linear least squares problem.
Another variant is to couple the curve fitting step with the gradient matching step, resulting both in higher accuracy and higher computational cost \citep{Ramsay2007}.

\paragraph{Gaussian process regression} The effort to formulate gradient matching probabilistically was spear-headed by \citet{Calderhead2009}, where Gaussian process regresion is combined with a product of experts approach.
This method was improved upon by \citet{Dondelinger2013} via joint sampling for GP and ODE parameters.
It was subsequently shown that a mean-field formulation can offer computational speed-ups \citep{Gorbach2017}.

\paragraph{In search for a generative model} There has been effort put to formulating Gaussian process-based gradient matching as inference in a generative model.
First by \citet{Barber2014}, who instead formulate a model directly linking state derivatives to measurements.
However, their approach suffers from identifiability problems, as demonstrated by \citet{Macdonald2015}.
It was later demonstrated by \citet{Wenk2019} that identifiability issues are also present for the product of experts approach.
They propose to resolve this issue by formulating an alternative model; this approach was pursued further by \citet{Wenk2020}.

\subsection{Probabilistic Numerics}
\paragraph{Relation to gradient matching} It might be tempting to interpret the probabilistic numerics approach as a variant of gradient matching.
But gradient matching fits a curve to the data and then the differential operator to the curve,
while for probabilistic numerics the order of operation is reversed:
\begin{enumerate}
\item Fit a curve by attempting to satisfy the differential equation at a finite set of points.
\item Fit the parameters of the differential operator by using the aforementioned curve and the data likelihood.
\end{enumerate}
The first step is implemented by probabilistic numerics, resulting in a \emph{physics-enhanced} Gaussian process prior, whereas the second step reduces to Gauss--Markov regression.
By directly incorporating the physics of the problem into the prior, it is ensured that inference is done in a well-posed probability model.
Consequently, issues regarding model specification and identifiability \citep{Macdonald2015,Wenk2019}, that have been recurring in gradient matching, are avoided.

\paragraph{Relation to numerical integration} The difference between probabilistic numerics and numerical integration for computing the likelihood comes down to the Dirac approximation $\widehat{\delta}_N$.
As can be seen in \eqref{eq:rk_likelihood}, numerical integration does so by simply approximating the locations of the Dirac.
On the other hand, probabilistic numerics approximates the Dirac with a distribution of non-zero width, often Gaussian in practice.
This has a smoothing effect on the likelihood and parallells can be drawn with the smoothing method in non-convex optimisation \citep{Mobahi2012}.
But the present method is not equivalent.
For example, the smoothing is not with respect to the variable of interest $\theta$, but rather with respect to the function $\varphi_{\theta}$.

\begin{figure*}[t]
  \centering
  \includegraphics[trim=0 4.7 0 4.3,clip]{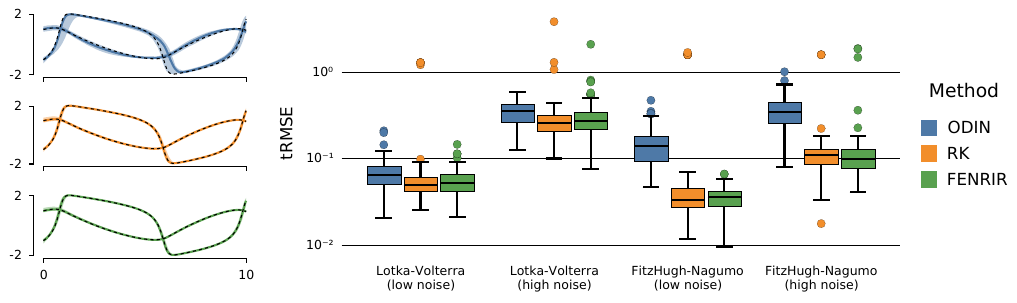}
  \caption{
    \textbf{Benchmarking estimation accuracy.}
    \emph{Left:} Trajectory summaries of 100 experiments, obtained by numerically integrating the inferred parameters of the FitzHugh--Nagumo system from noisy observations with high noise.
    The solid lines show the median trajectory, the shaded areas visualize the 10\% and 90\% quantiles, and the black dashed line shows the ground truth.
    \emph{Right:} Trajectory RMSEs (tRMSEs) on four benchmarks problems.
    Fenrir demonstrates performance that is competitive to ODIN and RK.
  }
  \label{fig:odin_comparison_trmses}
\end{figure*}

\paragraph{Previous probabilistic numerics approaches}
The probabilistic numerics approach to approximate the marginal likelihood has been explored to some extent by \citet{Kersting2020b}.
However, the present approach confers certain advantages over the former, the most notable being that the Gauss--Markov representation of the probabilistic solvers ensures all computations cost at most  $\mathrm{O}(N)$.

A probabilistic numerics approach has also been developed for estimating time varying parameters in the context of latent force modelling \citep{Schmidt2021}.
However, for the constant parameter problem, using linearised models can cause divergence in certain situations \citep{Ljung1979}.

An alternative to the inference-based methods hitherto discussed is to model the error by stochastic perturbation of numerical integrators \citep{Chkrebtii2016,Conrad2017,Matsuda2021,Teymur2018a}.

\section{Experimental Results}\label{sec:experiments}

This section investigates the utility and performance of Fenrir in a range of numerical experiments.
It is structured as follows.
\Cref{subsec:odin-experiments} evaluates Fenrir on two standard benchmark problems.
\Cref{subsec:model-selection} demonstrates the utility of the proposed marginal likelihood for model selection.
\Cref{subsec:partially-observed-states} considers systems with only partially observable states and shows that Fenrir, unlike most gradient matching methods, is still applicable.
Finally, \cref{subsec:high-oscillations} investigates highly oscillatory systems which present a particular challenge for numerical integration-based methods.

\paragraph{Implementation}
The implementation of the probabilistic numerical IVP solvers follows a number of practices for numerically stable implementation established by
\citet{Kramer2020a}.
All experiments are implemented in the Julia programming language
\citep{Bezanson_Julia_A_fresh_2017}.
Runge--Kutta reference solutions are computed with DifferentialEquations.jl
\citep{rackauckas2017differentialequations}, and
numerical optimizers are provided by Optim.jl
\citep{Optim.jl-2018}.
All experiments run on a single, consumer-level CPU.
Code is publicly available on GitHub.%
\footnote{\url{https://github.com/nathanaelbosch/fenrir-experiments}}

\subsection{Parameter Inference from Fully Observed States}
\label{sec:ex1}
\label{subsec:odin-experiments}
This experiment evaluates Fenrir on two benchmark problems that have been extensively studied in the both the gradient matching and the numerical integration literature \citep{Calderhead2009,Wenk2020}, namely the Lotka--Volterra predator-prey model and the FitzHugh--Nagumo neuronal model.
Detailed system descriptions, along with the ground-truth parameters, initial
values, and the chosen observation noise levels, are provided in \cref{app:subsec:5.1}.
We perform 100 experiments for each experimental setup, in which noisy observations are drawn from the numerically computed, true system trajectories.
The inference task then consists in estimating initial values and parameters from noisy state observations.
The quality of the resulting parameter estimates is evaluated using the
trajectory RMSE (tRMSE) metric as defined in \cref{def:trmse}.

We compare Fenrir to the probabilistic gradient matching method ODIN
\citep{Wenk2020} and to a non-linear least squares regression using a
Runge--Kutta solver, referred to as RK \citep{Bard1974}.
ODIN results are computed using the code published by \citet{Wenk2020};
RK is described in more detail in \cref{app:subsec:rk}.
All methods optimise their respective objectives with the L-BFGS algorithm
\citep{NoceWrig06}.
More details are provided in \cref{app:subsec:5.1}.

Results of the experiment are shown in \cref{fig:odin_comparison_trmses}.
In the median, Fenrir performs on par with ODIN and RK on Lotka--Volterra, but both RK and Fenrir outperform ODIN on FitzHugh--Nagumo and achieve more accurate state estimates as well as lower trajectory RMSEs.
Both RK and Fenrir suffer from outliers, but this issue appears to be less severe for Fenrir; see also
\cref{fig:ex1:absolute-errors} in \cref{app:subsec:5.1}.

\subsection{Model Selection}
\label{sec:ex2}
\label{subsec:model-selection}
For a given set of noisy observations, the true parametric form of the underlying system is often not known exactly.
Instead, a set of plausible models has to be evaluated against the observed data in order to find the most fitting candidate.
It has been previously shown that probabilistic gradient matching can be used for model selection, by comparing estimated noise parameters which are supposed to account for model mismatch \citep{Wenk2020}.
However, as Fenrir operates on a physics-informed probability model, model selection can be accomplished by statistically rigorous methods such as likelihood ratio testing \citep{King1998}.

The experiment follows the setup proposed by \citet{Wenk2020}.
We consider the Lotka--Volterra system as ground truth from which we numerically simulate experimental data, and generate a set of four candidate models by combining the true ODEs with two additional, incorrect equations -- all equations and parameters are provided in \cref{app:subsec:5.2}.
We obtain four models, \(\{M_{11}, M_{10}, M_{01}, M_{00}\}\), where \(M_{11}\) corresponds to the true Lotka--Volterra dynamics, \(M_{10}\) and \(M_{01}\) contain one correct and one wrong equation, and \(M_{00}\) contains only incorrect equations.
Thus, to succeed in this experiment, Fenrir should identify the correct model \(M_{11}\).

\begin{figure}[t]
  \centering
  \includegraphics{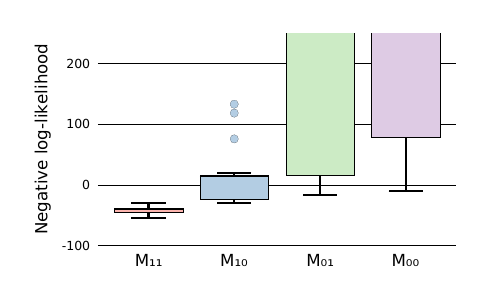}
  \caption{
    \textbf{Model selection results.}
    Fenrir correctly attributes the lowest negative log-likelihood (i.e. the highest probability) to the true \(M_{11}\) model.
    The figure is restricted to y-values up to 250 to show a clearer visualization, since the results vary largely in scale.
  }
  \label{fig:model_selection}
\end{figure}

We perform 100 individual model selection experiments to evaluate Fenrir's robustness regarding the observation noise.
The resulting marginal likelihoods are shown in \cref{fig:model_selection}.
We observe that Fenrir consistently attributes the lowest negative log-likelihood to the correct model \(M_{11}\), and is thus able to accurately identify the true model.

\begin{figure}[t]
  \centering
  \centering
  \includegraphics[]{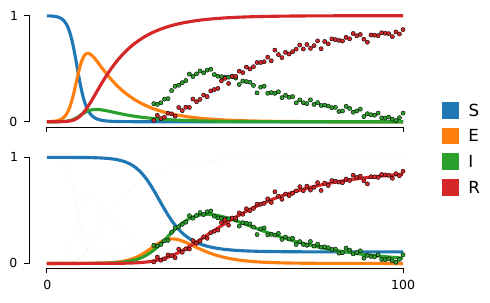}
  \caption{
    \textbf{Parameter inference in a SEIR model.}
    \emph{Top:} Trajectory resulting from the initial, randomly chosen parameters and initial values.
    \emph{Bottom:} Fenrir posterior after parameter optimization.
  }
  \label{fig:seir_inference_trajectory}
\end{figure}

\begin{figure}[t]
  \centering
  \includegraphics{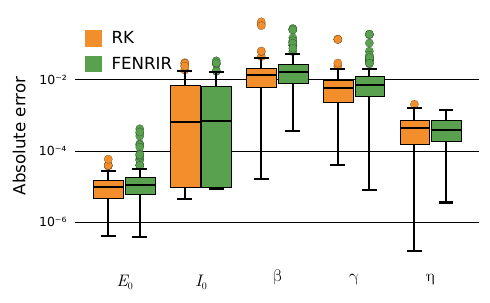}
  \caption{
    \textbf{Absolute parameter errors in the SEIR experiment.}
    Fenrir performs on par with the non-probabilistic Runge--Kutta baseline (RK) and is able to infer accurate parameter estimates from only partial observations of the SEIR system.
  }
  \label{fig:seir_parameter_errors}
\end{figure}

\subsection{Partially Observed System States}
\label{sec:ex3}
\label{subsec:partially-observed-states}

\begin{figure*}[t]
  \centering
  \includegraphics{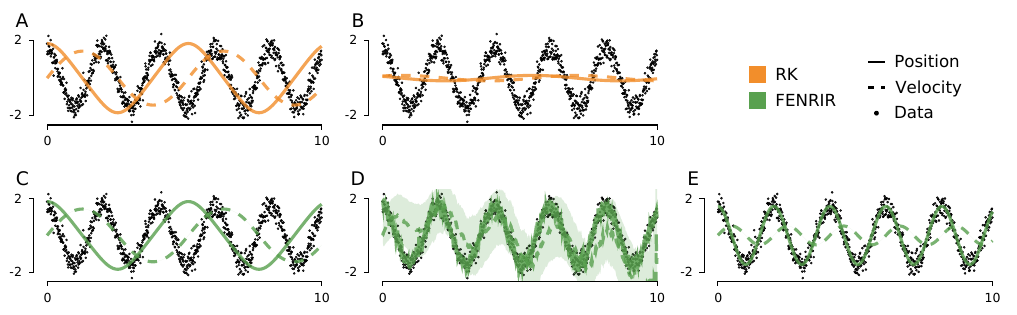}
  \caption{
    \textbf{Parameter inference in oscillatory systems.}
    Both RK and Fenrir start with an initial guess \(L_0=5.0\) for the pendulum length parameter [A,C].
    After optimization, the Runge--Kutta least-squares method RK ends up in a local minimum and is not able to recover the true parameter \(L^*=1.0\) [B].
    On the other hand, Fenrir first increases its diffusion hyperparameter to interpolate the data [D] (c.f.\ \cref{fig:pendulum_loss_landscape} below), and is then able to accurately recover the system parameter via optimization and provides accurate trajectory estimates [E].
  }
  \label{fig:pendulum_example}
\end{figure*}

Here, we evaluate Fenrir on an epidemeological model in which the system state can only be partially observed.
We consider a compartmental SEIR model that describes the fractions of a population that are
susceptible (S),
exposed (E),
infected (I; i.e.\ diagnosed with a positive test),
and recovered (R) over time
\citep{hethcote2000mathematics}.
Such compartmental models are commonly used to model the development of infectious diseases, and variants of the SEIR model have been used to explain COVID-19 outbreaks \citep{Menda2021}.
The definition of the dynamics, ground-truth initial values, and parameters are provided in \cref{app:subsec:5.3}.

At each point in time, only the infected and recovered population can be (approximately) observed, but the exposed and susceptible population is unknown.
Since Fenrir's ``dynamics-first'' approach only requires the observation to be linearly dependent on the system states (see \cref{eq:measurements,eq:measurements2}), no particular adjustments are needed for this experiment. 
Similarly, the Runge--Kutta-based approach considered in \cref{subsec:odin-experiments} is also applicable and will be used for comparison.
However, most gradient matching methods require all dimensions of the system states to be measurable in order to construct an interpolant, and are therefore not applicable to problems with partial observability.

\Cref{fig:seir_inference_trajectory} visualizes an individual experiment:
The initial values, parameters, and true system trajectories have to be estimated from noisy case counts of the infected and recovered population, which are furthermore given only from day \(30\) onwards.
The results of 100 experiments are shown in \cref{fig:seir_parameter_errors}.
Fenrir is able to consistently infer accurate parameter and trajectory estimates from noisy, partial observations of the dynamical system. 

\begin{figure}[t]
  \centering
  \includegraphics{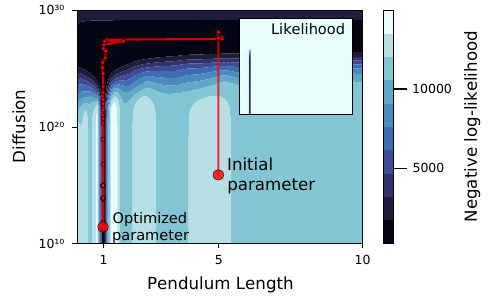}
  \caption{
    \textbf{Negative log-likelihood and optimization trajectory.}
    By first increasing its diffusion parameter, Fenrir
    is able to recover the true pendulum length parameter \(L=1\) by minimising the negative log-likelihood using L-BFGS.
    The likelihood (i.e.\ the negative exponential of the main plot) is shown in the inset figure.
  }
  \label{fig:pendulum_loss_landscape}
\end{figure}

\begin{figure}[t]
  \centering
  \includegraphics{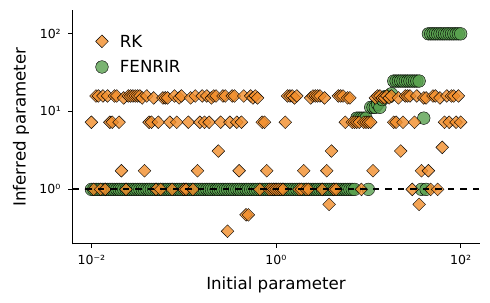}
  \caption{
    \textbf{Inferred parameters for various starting values.}
    Both RK and Fenrir are evaluated on a wide range of initial parameter estimates, from which they attempt to recover the true parameter \(L=1\) (dashed line) by optimization via L-BFGS.
    RK is often unable to approximate the true parameter,
    whereas Fenrir accurately recovers the true parameter for a wide range of starting points.
  }
  \label{fig:pendulum_optims}
\end{figure}

\subsection{Dynamical Systems with Fast Oscillations}
\label{sec:ex4}
\label{subsec:high-oscillations}
Finally, we evaluate Fenrir on a partially observable pendulum system that exhibits fast oscillations.
Problems of this form are known to be challenging for simulation-based methods such as the previously considered Runge--Kutta least-squares approach which, with poor initialization, often fail to capture the high frequencies
\citep{Benson1979}.
While gradient-matching methods are expected to be more robust to such problems, they require fully observable states and are therefore not applicable in the present setting.
Thus, we investigate Fenrir's capabilities of performing trajectory, parameter, and initial value inference under these challenges.

\Cref{fig:pendulum_example} visualizes the problem setup and a single experiment; a detailed description of the dynamics and the chosen hyperparameters is provided in \cref{app:subsec:5.4}.
In the shown example, the non-linear least squares regression converges towards the constant zero function and is unable to capture the high frequencies of the data.
On the other hand, by first optimizing the diffusion and observation noise parameters separately, Fenrir interpolates the experimental data and is then able to accurately approximate the true system parameters.
The chosen optimization trajectory is visualised with the corresponding loss landscape in \cref{fig:pendulum_loss_landscape}.
\Cref{fig:pendulum_optims} shows inferred parameters for a wider range of starting values; for simplicity, the initial value \(y_0\) is assumed to be known here.
RK often fails to converge towards the ground-truth, whereas Fenrir is able to recover the true parameter for a wide range of starting values.

\section{Conclusion}\label{sec:conclusion}
It has been demonstrated that the solution of an initial value problem can be approximated by a Gauss--Markov process, reducing the inference problem to Gauss--Markov regression.
The method offers advantages such as $\mathrm{O}(N)$ cost for inference, operability in the face of partial observations, regularised likelihoods, and moderate improvements in terms of estimation accuracy.
But, perhaps more importantly, it has been shown that probabilistic numerics is a promising method for rigorously incorporating physics in Gaussian process regression.


\section*{Acknowledgements}
The authors gratefully acknowledge financial support
by the German Federal Ministry of Education and
Research (BMBF) through Project ADIMEM (FKZ
01IS18052B), and financial support by the European
Research Council through ERC StG Action 757275 /
PANAMA; the DFG Cluster of Excellence “Machine
Learning - New Perspectives for Science”, EXC 2064/1,
project number 390727645; the German Federal Ministry of Education and Research (BMBF) through the
T¨ubingen AI Center (FKZ: 01IS18039A); and funds
from the Ministry of Science, Research and Arts of
the State of Baden-W\"urttemberg. The authors also
thank the International Max Planck Research School
for Intelligent Systems (IMPRS-IS) for supporting N.
Bosch.

\bibliography{../bib/refs}
\bibliographystyle{icml2022}

\newpage
\appendix
\onecolumn

\section{Additional Details on Probabilistic Numerics}\label{app:pn}
In this appendix, the probabilistic solver is described in detail.
Further details on the prior are given in \cref{app:pnprior}.
In \cref{app:pnlin}, it is explained how to compute the marginal moments and the parameters of the backward Markov representation of the posterior when the vector field is linear (affine).
In \cref{app:pnnonlin} some linearisation methods for approximate inference when the vector field is non-linear are reviewed.

\subsection{More details on priors}\label{app:pnprior}
Recall that the prior in state-space form is given by
\begin{subequations}
\begin{align}
\dif x(t) &= A x(t) \dif t + \sqrt{\kappa} B \dif w(t), \quad x(0) = x^\dagger_\theta,\\
y^{(m)}(t) &= \mathrm{E}_m^\T x(t), \quad m = 0,1,\ldots,\nu,
\end{align}
\end{subequations}
where $y^{(m)}$ models the $m$th derivative of the solution. By It\^o's formula this implies that
\begin{equation}
\dif \mathrm{E}_m^\T x(t) = \mathrm{E}_m^\T A x(t) \dif t + \sqrt{\kappa} \mathrm{E}_m^\T B \dif w(t),
\end{equation}
and for this to be consistent with the asserted derivative relations it must hold that
\begin{equation}
\mathrm{E}_m^\T A x(t) \dif t + \sqrt{\kappa} \mathrm{E}_m^\T B \dif w(t) = \mathrm{E}_{m+1}^\T x(t) \dif t, \quad m = 0,1,\ldots,\nu-1.
\end{equation}
This in turn implies that it must hold that
\begin{subequations}
\begin{align}
\mathrm{E}_m^\T A  &= \mathrm{E}_{m+1}^\T, \quad m= 0,1,\ldots,\nu-1, \\
\mathrm{E}_m^\T B  &= 0, \quad m=0,1,\ldots,\nu-1,
\end{align}
\end{subequations}
while $\mathrm{E}_\nu^\T A$ and $\mathrm{E}_\nu^\T B$ are free parameters.
Letting $\mathrm{e}_m$ be the $m$th canonical basis vector in $\mathbb{R}^{\nu+1}$, $\mathrm{I}_d$ be the $d$ by $d$ identity matrix, and fixing  $\mathrm{E}_m = \mathrm{e}_m \otimes \mathrm{I}_d$ then gives the model
\begin{equation}
\dif y^{(\nu)}(t) = \sum_{m=0}^\nu A_{\nu,m} y^{(m)} \dif t + \sqrt{\kappa} B_\nu \dif w(t),
\end{equation}
where $A_{\nu,m} = \mathrm{E}_\nu^\T A \mathrm{E}_m$ and $B_\nu = \mathrm{E}_\nu^\T B$.
Any other state-space model of dimension $d(\nu+1)$  modelling a vector valued function of dimension $d$ and its $\nu$ first derivatives must be related to this via similarity transform.
The canonical model in probabilistic numerics is the $\nu$-times integrated Wiener process \citep{Schober2019,Tronarp2019c,Kramer2020a,Bosch2021,Kersting2020a}, where the parameters are given by
\begin{equation}
A_{\nu,m} = 0, \quad m=0,1,\ldots,\nu-1.
\end{equation}
Though other priors are of course possible \citep{Magnani2017,Tronarp2021a,Kersting2020a}. Usually, the diffusion matrix $B_\nu$ is set to identity as well, yielding the following prior
\begin{equation}
\dif y^{(\nu)}(t) =  \sqrt{\kappa} \dif w(t),
\end{equation}
which is used throughout the article.

\subsection{Posterior for linear vector fields}\label{app:pnlin}
Suppose the vector field is linear:
\begin{equation}
f_\theta(t,y) = L_\theta(t) y + b_\theta(t),
\end{equation}
then the probabilistic IVP solver reduces to inference in the following model:
\begin{equation}
\dif x(t) = A x(t) \dif t + \sqrt{\kappa} B \dif w(t), \quad x(0) = x^\dagger_\theta,\\
\end{equation}
subject to the data
\begin{subequations}
\begin{align}
C_\theta(t) &= \mathrm{E}_1 - \mathrm{E}_0 L_\theta^\T(t), \\
z(t) &= 0 = \mathrm{E}_1^\T x(t)   - L_\theta(t) \mathrm{E}_0^\T x(t) - b_\theta(t)  , \quad t \in \mathbb{T}_{ \mathsf{PN}}.
\end{align}
\end{subequations}
The posterior is Gaussian because the prior is Gaussian and the measurement functionals are linear, and it can be computed with the well-known forward / backward recursions \citep{Kalman1960,RauchTungStriebel1965}.

More specifically, denote the numerics data up to time $t$ by
\begin{equation}
\mathscr{Z}_{[0,t]} = \big \{ z(t) = 0 \colon t \in \mathbb{T}_{\mathsf{PN}} \cap [0,t]   \big \}
\end{equation}
and up to \emph{just before} time $t$ by
\begin{equation}
\mathscr{Z}_{[0,t)} = \big \{ z(t) = 0 \colon t \in \mathbb{T}_{\mathsf{PN}} \cap [0,t)   \big \}.
\end{equation}
The forward recursion then computes the filtering densities
\begin{equation}
p(t,x \mid \mathscr{Z}_{[0,t]}) = \mathcal{N}(x; \mu_\theta(t), \kappa \Sigma_\theta(t)),
\end{equation}
which agree with the prediction densities
\begin{equation}
p(t,x \mid \mathscr{Z}_{[0,t)}) = \mathcal{N}(x; \mu_\theta(t^-), \kappa \Sigma_\theta(t^-)),
\end{equation}
unless $t \in \mathbb{T}_{\mathsf{PN}}$.
The filtering moments are then post-processed in the backwards recursion to produce the smoothing densities (time marginals of the posterior)
\begin{equation}
p(t,x \mid \mathscr{Z}(t_N)) = \mathcal{N}(x; \xi_\theta(t), \kappa \Lambda_\theta(t)).
\end{equation}
For a more thorough exposition on filtering and smoothing refer to \citet{Sarkka2013,Sarkka2019}.
Furthermore, the fact that the scaling $\kappa$ is retained throughout the recursion follows from the fact that the initial covariance and all transition covariances are scaled by $\kappa$ \citep{Tronarp2019d}.

\paragraph{Forward recursion} The forward recursion starts by initialising the filter mean and covariance according to
\begin{subequations}
\begin{align}
\mu_\theta(t_0) &= x^\dagger_\theta, \\
\Sigma_\theta(t_0) &= 0,
\end{align}
\end{subequations}
whereafter the algorithm alternates between prediction and update. The prediction equations are given by
\begin{subequations}
\begin{align}
\mu_\theta(t_n^-)     &= \Phi(\Delta_n) \mu_\theta(t_{n-1}), \\
\Sigma_\theta(t_n^-)  &= \Phi(\Delta_n) \Sigma_\theta(t_{n-1}) \Phi^\T(\Delta_n) +  Q(\Delta_n), \\
G_\theta(t_{n-1})     &= \Sigma_\theta(t_{n-1}) \Phi^\T(\Delta_n)\Sigma_\theta^{-1}(t_n^-), \\
P_\theta(t_{n-1})     &= \Sigma_\theta(t_{n-1}) - G_\theta(t_{n-1}) \Sigma_\theta(t_n^-) G_\theta^\T(t_{n-1}),
\end{align}
\end{subequations}
where $G_\theta$ and $P_\theta$ are parameters associated with the subsequent backward recursion. The update relations are given by
\begin{subequations}
\begin{align}
C_\theta(t_n) &= \mathrm{E}_1 - \mathrm{E}_0 L_\theta^\T(t_n), \\
S_\theta(t_n) &= C_\theta^\T(t_n) \Sigma_\theta(t_n^-) C_\theta(t_n), \\
K_\theta(t_n) &= \Sigma_\theta(t_n^-) C_\theta^\T(t_n) S_\theta^{-1}(t_n), \\
\mu_\theta(t_n) &= \mu_\theta(t_n^-) + K_\theta(t_n)\big( b_\theta(t_n) - C_\theta^\T(t_n) \mu_\theta(t_n^-)  \big),\\
\Sigma_\theta(t_n) &= \Sigma_\theta(t_n^-) - K_\theta(t_n) S_\theta(t_n) K_\theta^\T(t_n).
\end{align}
\end{subequations}

\paragraph{Backward recursion} The backwards recursion starts by setting the smoother mean and covariance to the filter mean and covariance at the terminal point according to
\begin{subequations}
\begin{align}
\xi_\theta(t_N) &= \mu_\theta(t_N), \\
\Lambda_\theta(t_N) &= \Sigma_\theta(t_N).
\end{align}
\end{subequations}
The backwards recursion is then given by
\begin{subequations}
\begin{align}
\xi_\theta(t_n)  &= \mu_\theta(t_n) +  G_\theta(t_n) \big( \xi_\theta(t_{n+1})  - \Phi(\Delta_{n+1}) \mu_\theta(t_n)     \big), \\
\Lambda_\theta(t_n) &= G_\theta(t_n) \Lambda_\theta(t_{n+1}) G_\theta^\T(t_n) + P_\theta(t_n).
\end{align}
\end{subequations}

\paragraph{Backward Markov process representation} Lastly, the posterior may be represented, on the grid, by the following backwards Markov process
\begin{equation}
\begin{split}
&\gamma_N(x(t_{1:N})\mid \theta, \kappa) = \mathcal{N} \big( x(t_N) ; \xi_\theta(t_N), \kappa \Lambda_\theta(t_N)   \big) \\
&\quad \prod_{n=N-1}^1 \mathcal{N}\big( x(t_n) ; \mu_\theta(t_n) + G_\theta(t_n) \big( x(t_{n+1}) - \mu_\theta(t_{n+1}^-) \big), \kappa P_\theta(t_n)   \big).
\end{split}
\end{equation}
This follows from the fact that
\begin{equation}
p(t,x \mid s, x', \mathscr{Z}_{[0,T]}) = p(t,x \mid s, x', \mathscr{Z}_{[0,t]}), \quad t_{n+1} \geq s > t \geq t_n, \ n=1,\ldots,N.
\end{equation}
That is, by total probability
\begin{equation}
\begin{split}
p(t_n,x_n\mid \mathscr{Z}_{[0,T]}) &= \int p(t_n,x_n\mid t_{n+1},x_{n+1}, \mathscr{Z}_{[0,T]}) p(t_{n+1},x_{n+1}\mid \mathscr{Z}_{[0,T]}) \dif x_{n+1} \\
&= \int p(t_n,x_n\mid t_{n+1},x_{n+1}, \mathscr{Z}_{[0,t_n]}) p(t_{n+1},x_{n+1}\mid \mathscr{Z}_{[0,T]}) \dif x_{n+1},
\end{split}
\end{equation}
and by Bayes' rule
\begin{equation}
\begin{split}
p(t_n,x_n\mid t_{n+1},x_{n+1}, \mathscr{Z}_{[0,t_n]}) &\propto p(t_n,x_n \mid \mathscr{Z}_{0,t_n}) p(t_{n+1},x_{n+1} \mid t_n, x_n, \mathscr{Z}_{[0,t_n]}) \\
&= \mathcal{N}( x_n; \mu_\theta(t_n), \kappa\Sigma_\theta(t_n)) \mathcal{N}(x_{n+1}; \Phi(\Delta_{n+1})x_n, \kappa Q(\Delta_{n+1})) \\
&= \mathcal{N}(x_{n+1}; \mu_\theta(t_{n+1}^-),\Sigma_\theta(t_{n+1}^-)) \mathcal{N}(x_n; \mu_\theta(t_n) + G_\theta(t_n)( x_{n+1} - \mu_\theta(t_{n+1}^-)  ), \kappa P_\theta(t_n) ) \\
&\propto \mathcal{N}(x_n; \mu_\theta(t_n) + G_\theta(t_n)( x_{n+1} - \mu_\theta(t_{n+1}^-)  ), \kappa P_\theta(t_n) ), \\
\end{split}
\end{equation}
where the last equality follows from ordinary Gaussian conditioning and the proportionality signs are with respect to $x_n$.
This proves the recursive structure of the posterior as asserted by proposition \ref{prop:gauss_markov_pn_post},
and the complete result follows from the fact that the marginal filtering and smoothing densities coincide at the terminal point.
That is,
\begin{equation}
p(t_N,x_N,\mid \mathscr{Z}_{[0,T]}) = \mathcal{N}( x_N ; \mu_\theta(t_N), \kappa \Sigma_\theta(t_N) ) = \mathcal{N}(x_N; \xi_\theta(t_N), \kappa \Lambda_\theta(t_N)).
\end{equation}

\subsection{Approximate posteriors via linearisation}\label{app:pnnonlin}
When the vector field is non-linear, the posterior is in most cases intractable.
However, approximate posteriors may be obtained by linearising the data relation in \eqref{eq:pn_data}.
Due to the structure of the information operator, there are multiple choices for doing this, namely
\begin{enumerate}
\item Zeroth order linearisation \citep{Schober2019}:
\begin{subequations}
\begin{align}
\hat{L}_\theta(t) &= 0, \\
\hat{b}_\theta(t) &= f_\theta(t,\tilde{y}(t))
\end{align}
\end{subequations}
\item First order linearisation \citep{Tronarp2019c}:
\begin{subequations}
\begin{align}
\hat{L}_\theta(t) &= J_{f_\theta}(t,\tilde{y}(t)), \\
\hat{b}_\theta(t) &=  f_\theta(t,\tilde{y}(t)) - J_{f_\theta}(t,\tilde{y}(t)) \tilde{y}(t)
\end{align}
\end{subequations}
\end{enumerate}
The linearisation point is typicaly chosen as the predictive mean:
\begin{equation}
\tilde{y}(t) = \mathrm{E}_0^\T \mu_\theta(t^-).
\end{equation}
However, other choices are possible as well, such as the smoothing mean \citep{Tronarp2021a}
\begin{equation}
\tilde{y}(t) = \mathrm{E}_0^\T \xi_\theta(t),
\end{equation}
which leads to the fixed-point equations for the Gauss--Newton algorithm \citep{Bell1994}.

\section{Inference in IVPs as Gauss--Markov regression}\label{app:ivp2gmr}
Using the probabilistic numerics posterior as a surrogate for the solution of the initial value problem leads to the following inference problem
\begin{subequations}
\begin{align}
x(t_N) &\sim \mathcal{N}( \xi_\theta(t_N), \kappa \Lambda_\theta(t_N) ), \\
x(t_n) \mid x(t_{n+1}) &\sim \widehat{\gamma}_N(x(t_n) \mid x(t_{n+1}) \mid \theta, \kappa), \\
u(t) \mid x(t) &\sim \mathcal{N}(  H^\T \mathrm{E}_0^\T x(t_n), R_\theta ), \quad t \in \mathbb{T}_{\mathsf{D}}.
\end{align}
\end{subequations}
This is again, a problem of Gauss--Markov regresion and can be solved by the usual forward / backward recursions.
What is unusual is that the latent process is specified in terms of a terminal distribution and backward transition densities. Therefore, the equations required for implementation are given in detail.

\subsection{The forward (but backward in time) recursion and the marginal likelihood}
The backward recursion is implemented by a forward recursion with flipped time axis. That is, start by initialising the filter moments:
\begin{subequations}
\begin{align}
\breve{\mu}_\theta(t_N^+)    &= \xi_\theta(t_N), \\
\breve{\Sigma}_\theta(t_N^+) &= \kappa \Lambda_\theta(t_N),
\end{align}
\end{subequations}
whereafter the algorithm alternates between a backward prediction and update. If $t_n \in \mathbb{T}_{\mathsf{D}}$, then an update is performed according to
\begin{subequations}
\begin{align}
\breve{H}           &= \mathrm{E}_0 H, \\
\breve{S}(t_n)      &= \breve{H}^\T \breve{\Sigma}_\theta(t_n) \breve{H} + R_\theta, \\
\breve{K}_\theta(t_n)      &= \breve{\Sigma}_\theta(t_n) \breve{H} \breve{S}_\theta^{-1}(t_n), \\
\breve{\mu}_\theta(t_n)    &= \breve{\mu}_\theta(t_n^+) + \breve{K}_\theta(t_n) \big( u(t_n) -   \breve{H}^\T \breve{\mu}_\theta(t_n^+)  \big), \\
\breve{\Sigma}_\theta(t_n) &= \breve{\Sigma}_\theta(t_n^+) - \breve{K}_\theta(t_n) \breve{S}_\theta(t_n) \breve{K}_\theta^\T(t_n).
\end{align}
\end{subequations}
The prediction step is given by
\begin{subequations}
\begin{align}
\breve{\mu}_\theta(t_{n-1}^+) &= \mu_\theta(t_{n-1}) + G_\theta(t_{n-1}) \big( \breve{\mu}_\theta(t_{n+1}) - \mu_\theta(t_{n+1}^-) \big),\\
\breve{\Sigma}_\theta(t_{n-1}^+) &= G_\theta(t_{n-1})  \breve{\Sigma}_\theta(t_n) G_\theta^\T(t_{n-1}) + \kappa P_\theta(t_{n-1}).
\end{align}
\end{subequations}
Finally, the marginal likelihood approximation is given by the prediction error decomposition \citep{Schweppe1965}
\begin{equation}
\widehat{\mathcal{M}}_N(\theta,\kappa) = \prod_{t \in \mathbb{T}_{\mathsf{D}} } \mathcal{N}\big( u(t) ;  \breve{H}^\T \breve{\mu}_\theta(t^+), \breve{S}_\theta(t)    \big).
\end{equation}

\subsection{The backward (but forward in time) recursion and trajectory estimates}
The smoothing parameters for the forward recursion are given by
\begin{subequations}
\begin{align}
\breve{G}_\theta(t_n) &=  \breve{\Sigma}_\theta(t_n) G_\theta^\T(t_{n-1}) \breve{\Sigma}_\theta^{-1}(t_{n-1}^+) ,\\
\breve{P}_\theta(t_n)         &=  \breve{\Sigma}_\theta(t_n) - \breve{G}_\theta(t_n) \breve{\Sigma}_\theta(t_{n-1}^+) \breve{G}_\theta^\T(t_n),
\end{align}
\end{subequations}
and the forward smoothing recursion is given by
\begin{subequations}
\begin{align}
\breve{\xi}_\theta(t_n)  &= \breve{\mu}_\theta(t_n)  + \breve{G}_\theta(t_n) \big(  \breve{\xi}_\theta(t_{n-1})  -     G_\theta(t_{n-1}) \breve{\mu}_\theta(t_n)  \big), \\
\breve{\Lambda}_\theta(t_n) &=  \breve{G}_\theta(t_n) \breve{\Lambda}_\theta(t_{n-1})  \breve{G}_\theta^\T(t_n) + \breve{P}_\theta(t_n).
\end{align}
\end{subequations}

\section{Additional Details on the Experimental Evaluation}

In all experiments, Fenrir uses a 5-times integrated Wiener process prior and a first-order linearisation of the vector field during the probabilistic numerical ODE solve when computing its physics-enhanced prior.

\paragraph{Optimization}
Throughout all experiments, the L-BFGS method
has been used for optimization with both Fenrir and RK
\citep{NoceWrig06};
L-BFGS is also the optimizer of choice in the official ODIN code by \citet{Wenk2020}.
The specific L-BFGS implementation is provided by the Optim.jl software package
\citep{Optim.jl-2018}.
In all experiment, the observation noise \(\sigma^2\) and the diffusion \(\kappa\) are optimised in log-space.

\paragraph{Parameter Initialization}
As done in ODIN, ODE parameters are initialised with a folded normal distribution, i.e. as the absolute value of a sample from standard normal Gaussian, and initial values are initialised with their noisy observation \(u(t_0)\), unless specified otherwise.
Observation noise is always initialised as \(\sigma^2=1\).

\subsection{Baseline: Non-linear Least Squares Regression using a Runge--Kutta Solver}
\label{app:subsec:rk}
Given data \(\mathcal{D} = \{u(t)\}\) on the grid \(t \in \mathbb{T}_{\mathsf{D}}\), the considered ``RK'' baseline method minimizes the loss
\begin{equation}
  L := \sum_{t \in \mathbb{T}_\mathsf{D}} \left\| H \cdot \hat{y}(t) - u(t) \right\|_2^2,
\end{equation}
where \(\hat{y}(t)\) is computed with a classical Runge--Kutta initial value solver and \(H\) is the measurement matrix as introduced in \cref{eq:measurements}.
In most experiments, the
\texttt{Tsit5} \citep{tsit5} solver is used, with adaptive step-size selection for absolute and relative tolerances \(\tau_\text{abs} = 10^{-8}\), \(\tau_\text{rel} = 10^{-6}\).
Only on the FitzHugh-Nagumo system we use the implicit \texttt{RadauIIA5} \citep{radau} method, since we observed it to be more robust as some parameter settings can lead to stiff dynamics.
Both solvers are provided by DifferentialEquations.jl
\citep{rackauckas2017differentialequations}.

\begin{figure*}[p]
  \centering
  \begin{subfigure}{\textwidth}
    \centering
  \includegraphics{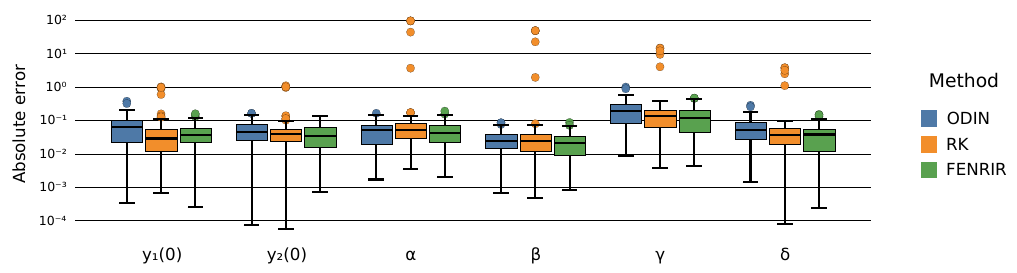}
  \caption{Lotka--Volterra with low observation noise.}
\end{subfigure}

\begin{subfigure}{\textwidth}
  \centering
  \includegraphics{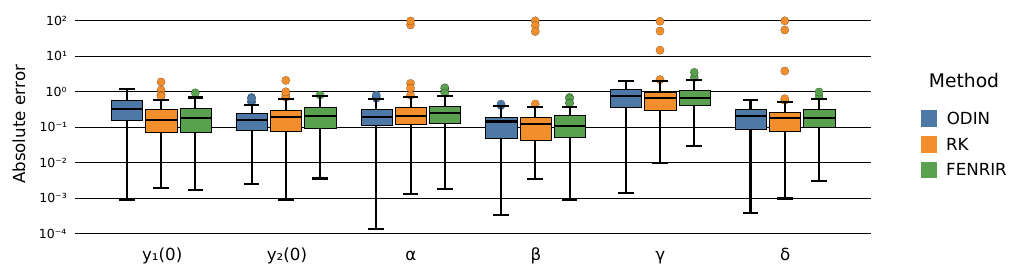}
  \caption{Lotka--Volterra with high observation noise.}
\end{subfigure}

\begin{subfigure}{\textwidth}
  \centering
  \includegraphics{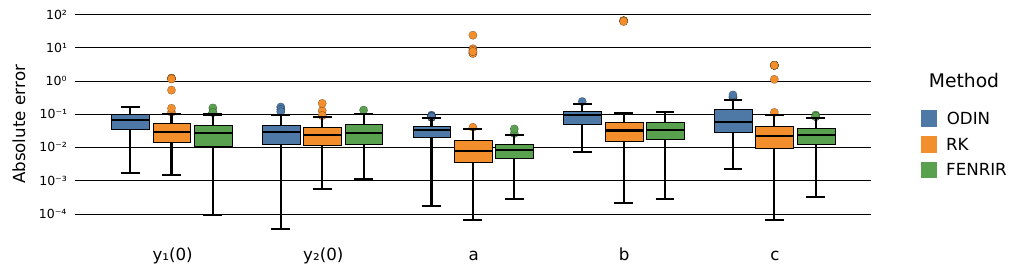}
  \caption{FitzHugh--Nagumo with low observation noise.}
\end{subfigure}

\begin{subfigure}{\textwidth}
  \centering
  \includegraphics{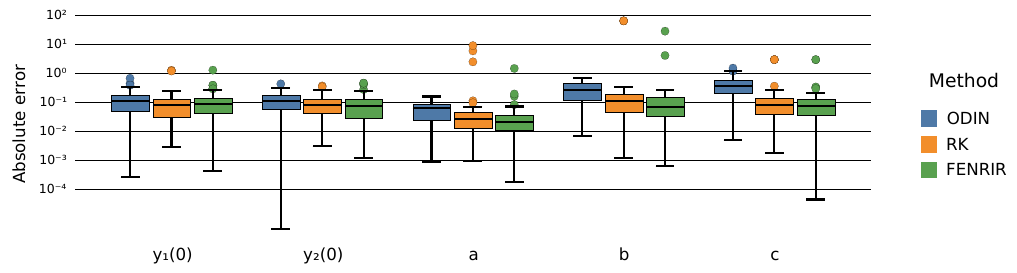}
  \caption{FitzHugh--Nagumo with high observation noise.}
\end{subfigure}
\caption{\textbf{Absolute parameter errors.}}
\label{fig:ex1:absolute-errors}
\end{figure*}

\subsection{Additional Details on \cref{sec:ex1}: ``Parameter Inference from Fully Observed States''}
\label{app:subsec:5.1}

\begin{definition}[Trajectory RMSE]
  \label{def:trmse}
  Let \(\hat{\theta}\) be the parameters estimated by an inference algorithm, and let \(\mathbb{T}_\mathsf{D}\) be the set of measurement nodes.
  Then, let \(\hat{y}(t)\), \(t \in \mathbb{T}_\mathsf{D}\), be the estimated system trajectory, computed by numerically integrating the ODE with initial values and parameters as given by the estimated \(\hat{\theta}\).
  The trajectory RMSE (tRMSE) is then defined as
  \begin{equation}
    \text{tRMSE} := \sqrt{\frac{1}{\left| \mathbb{T}_\mathsf{D} \right|} \sum_{t \in \mathbb{T}_\mathsf{D}} \left\| \hat{y}(t) - y(t) \right\|_2^2}.
  \end{equation}
\end{definition}

\paragraph{Lotka--Volterra}
The Lotka--Volterra model describes the dynamics of biological systems in which two species interact, one as a predator and the other as prey
\citep{lotka,volterra}.
It is described by the ODEs
\begin{subequations}
  \begin{align}
    \dot{y_1} &= \alpha y_1 - \beta y_1 y_2, \label{eq:modelselection:i1} \\
    \dot{y_2} &= -\gamma y_1 + \delta y_1 y_2. \label{eq:modelselection:j1}
  \end{align}
\end{subequations}
As ground truth, we assume an initial value \(y_0 = [5,3]^\T\) and parameters
\(\alpha=2\), \(\beta=1\), \(\gamma=4\), \(\delta=1\).
The experimental data is generated on the equi-spaced time grid
\(t_i \in \mathbb{T}_\mathsf{D} = \{0.0, 0.1, \dots, 2.0\}\), as
\(u(t_i) = \hat{y}(t_i) + v(t_i)\),
where \(\hat{y}(t_i)\) is computed via accurate, numerical simulation, and with noise
\(v(t) \sim \mathcal{N}(0, \sigma^2 \cdot I)\).
We further consider two different noise levels \(\sigma^2_\text{low}=0.01\) and \(\sigma^2_\text{high}=0.25\).
Thus, the full set of parameters to be estimated is \(\theta = \{y_0, \alpha, \beta, \gamma, \delta, \sigma\}\), as well as the diffusion \(\kappa\).
In this system, we found it helpful to first optimize the noise and diffusion parameters \(\sigma, \kappa\) individually until convergence, and only then optimize all parameters jointly;
such an approach is also chosen by the gradient matching method ODIN \citep{Wenk2020}.
Furthermore, as in the original experimental setup by \citet{Wenk2020}, we consider bounds \(y_0 \in [0,100]^2\), \(\alpha,\beta,\gamma,\delta \in [0,100]\), \(\sigma^2 \in [10^{-6},10^{2}]\), and additionally \(\kappa \in [10^{-20}, 10^{50}]\).
Finally, a step-size of \(\Delta = 5 \cdot 10^{-3}\) is chosen for Fenrir's probabilistic numerical integration.

\paragraph{FitzHugh--Nagumo}
The FitzHugh--Nagumo neuronal model
\citep{FitzHugh1955,Nagumo1962}
is given by the ODE
\begin{subequations}
  \begin{align}
    \dot{y}_1 &= c \left( y_1 - \frac{y_1^3}{3} + y_2 \right), \\
    \dot{y}_2 &= -\frac{1}{c} \left( y_1 -  a - b y_2 \right).
  \end{align}
\end{subequations}
We consider ground-truth parameters
\(a=0.2\), \(b=0.2\), \(c=3.0\),
and a true initial value
\(y_0 = [-1, 1]^\T\).
The experimental data is generated on the grid
\(t_i \in \mathbb{T}_\mathsf{D} = \{0.0, 0.5, \dots, 10.0\}\),
by disturbing a high-confidence numerical simulation of the true trajectory with Gaussian noise
\(v(t) \sim \mathcal{N}(0, \sigma^2 \cdot I)\),
for two noise levels \(\sigma^2_\text{low}=0.005\) and \(\sigma^2_\text{high}=0.05\).
The full set of (hyper)parameters to be estimated by Fenrir is then \(\theta = \{y_0, \alpha, \beta, \gamma, \delta, \sigma\}\), as well as the diffusion \(\kappa\).
All of which are jointly optimised via L-BFGS, while assuming bounds \(y_0 \in [-100,100]^2\), \(a,b,c \in [0,100]\), \(\sigma^2 \in [10^{-6},10^{2}]\), and \(\kappa \in [10^{-20}, 10^{50}]\).
Fenrir's physics-enhanced prior is computed with a step size \(\Delta = 10^{-2}\).

\subsection{Additional Details on \cref{sec:ex2}: ``Model Selection''}
\label{app:subsec:5.2}
The Lotka--Volterra model with ground-truth parameters as described in \cref{app:subsec:5.1} is extended to a set of four candidate models, via the following additional ODEs:
\begin{subequations}
  \begin{align}
    \dot{y_1} &= \alpha y_1^2 - \beta y_2, \label{eq:modelselection:i0} \\
    \dot{y_2} &= -\gamma y_2. \label{eq:modelselection:j0}
  \end{align}
\end{subequations}
By combining these two wrong equations with the true ODEs, we obtain for models \(M_{ij}\), with \(i,j \in \{0,1\}\) indicating if the correct (1) or incorrect equation (0) has been used;
for instance, \(M_{01}\) contains \cref{eq:modelselection:i0} and \cref{eq:modelselection:j1}.
The experimental data is generated as described in \cref{app:subsec:5.1}, with a ``low'' noise setting of \(\sigma_\text{low}^2=0.01\).
All parameters are optimised jointly by Fenrir via L-BFGS, with bounds for parameters and initial values chosen as in \cref{app:subsec:5.1}.

\subsection{Additional Details on \cref{sec:ex3}: ``Partially Observed System States''}
\label{app:subsec:5.3}

The compartmental SEIR model \citep{hethcote2000mathematics}
describes the fractions of a population that are susceptible (S), exposed (E), infected (I; i.e. diagnosed by a positive test), and recovered (R).
It is given in as differential equations
\begin{subequations}
  \begin{align}
    \dot{S} &= - (\beta_E \cdot S \cdot E + \beta_I \cdot S \cdot I), \\
    \dot{E} &= \beta_E \cdot S \cdot E + \beta_I \cdot S \cdot I - \gamma \cdot E, \\
    \dot{I} &= \gamma \cdot E - \lambda \cdot I, \\
    \dot{R} &= \lambda \cdot I.
  \end{align}
\end{subequations}
with infection rates
\(\beta_E\) and \(\beta_I\),
transition rate \(\gamma\) from exposure to infection,
and recovery / death rate \(\lambda\).
Following \citet{Menda2021}, which used an extension of the SEIR model to explain COVID-19 outbreaks, we
consider ground-truth parameters
\(\beta_I=0\), \(\beta_E = 0.5\), \(\gamma = 1/5\), and \(\lambda = 1/21\)
(the latter two correspond to realistic estimates of transition and recovery rate in COVID-19, given by \citet{lauer2020incubation,Bi2020}).
Furthermore, we generate data on the time grid \(\mathbb{T}_\mathsf{D} = \{30, 31, \dots, 100\}\) from initial values
\(E_0=10^{-4}\), \(I_0=10^{-5}\), \(R_0=0\), and \(S_0=1-E_0-I_0\) at time \(t_0=0\),
as
\(u(t_i) = H \cdot \hat{y}(t_i) + v(t_i)\),
with a measurement matrix
\begin{equation}
  H =
  \begin{bmatrix}
    0 & 0 & 1 & 0 \\
    0 & 0 & 0 & 1
  \end{bmatrix}
\end{equation}
such that only the infected and recovered population is measured, and
disturbed by Gaussian noise
\(v(t) \sim \mathcal{N}(0, \sigma^2 \cdot I)\) with \(\sigma^2=5 \cdot 10^{-4}\).

Instead of estimating the full initial state, we parameterize it by the initial exposed and infected population count:
\begin{equation}
  y_0(E_0, I_0) = \begin{bmatrix}1 - E_0 - I_0,& E_0,& I_0,& 0 \end{bmatrix}^\T.
\end{equation}
Thus, the parameters to be estimated by Fenrir in this experiment are
\(\theta = \{E_0, I_0, \beta_E, \gamma, \lambda, \sigma\}\), as well as the diffusion \(\kappa\).
All parameters are jointly optimised via L-BFGS, with bounds
\(E_0,I_0,\beta_E,\gamma,\lambda \in [0,1]\), \(\sigma^2 \in [10^{-6},10^{2}]\), and \(\kappa \in [10^{-20}, 10^{20}]\).
In each experiment, ODE parameters \(\beta_E,\gamma,\lambda\) are initialised as uniformly random; the starting values for \(E_0, I_0\) are initialised as absolute values of samples from a Gaussian \(\mathcal{N}(0, 10^{-2})\).
Fenrir's probabilistic numerical integration is performed with a step size \(\Delta=0.2\).

\subsection{Additional Details on \cref{sec:ex4}: ``Dynamical Systems with Fast Oscillations''}
\label{app:subsec:5.4}
The considered pendulum system is given by a second-order ODE
\(\ddot{y}=-\frac{g}{L}\sin(y)\), which can be transformed to the following first-order equations
\begin{subequations}
  \begin{align}
    \dot{y_1} &= y_2, \\
    \dot{y_2} &= - \frac{g}{L} \sin(y_1),
  \end{align}
\end{subequations}
with the gravity constant \(g=9.81\).
We assume a ground-truth parameter \(L=1\) and an initial value \(y_0=[0, \pi/2]\).
The observation data is generated as
\(u(t_i) = \begin{bmatrix}0&1\end{bmatrix} \cdot \hat{y}(t_i) + v_i\), with
observation noise \(v(t_i) \sim \mathcal{N}(0, \sigma^2)\), \(\sigma^2=0.1\),
on the grid \(t_i \in \mathbb{T}_\mathsf{D} = \{0.01 \cdot i\}_{i=0}^{1000}\).
In the corresponding experiment, we found it to be beneficial to first optimize the noise \(\sigma\) and diffusion parameter \(\kappa\), before jointly optimizing all model parameters \(\theta = \{y_0, L, \sigma\}\) and the diffusion \(\kappa\).
while assuming bounds
\(y_0 \in [-100,100]^2\),
\(L \in [0,100]\),
\(\sigma^2 \in [10^{-8},10^{4}]\), and
\(\kappa \in [10^{-20}, 10^{50}]\).
Finally, Fenrir's physics-enhanced prior is computed with a fixed step size \(\Delta=0.1\).

\end{document}